\documentclass{article}

\usepackage[preprint]{neurips_2019}

\usepackage[utf8]{inputenc} 
\usepackage[T1]{fontenc}    
\usepackage[dvipsnames]{xcolor}
\usepackage[colorlinks=true,linkcolor=blue,citecolor=Gray]{hyperref}       
\usepackage{url}            
\usepackage{booktabs}       
\usepackage{amsfonts}       
\usepackage{nicefrac}       
\usepackage{microtype}      
\usepackage{amsmath}

\usepackage{graphicx}
\usepackage{graphicx}
\usepackage{caption,subcaption}

\newcommand{\htilde}{\widetilde{\textrm{h}}}
\newcommand{\normh}[2]{\htilde_{#1}^{(#2)}}

\newcommand{\resultFigs}[4]{
    \begin{figure}[ht]
    \begin{subfigure}[b]{0.5\linewidth}
        \centering
    \includegraphics[width=0.95\textwidth]{#2}
        \caption{}\label{fig:#1a}
    \end{subfigure}%
    \begin{subfigure}[b]{0.5\linewidth}
        \centering
    \includegraphics[width=0.45\textwidth]{example-image-a}
        \hfil
    \includegraphics[width=0.45\textwidth]{example-image-b}\\[2mm]
    \includegraphics[width=0.45\textwidth]{example-image-c}
        \hfil
    \includegraphics[width=0.45\textwidth]{example-image}
        \caption{}\label{fig:fig2}
      \end{subfigure}   
}

\title{\textsc{exBERT}: A Visual Analysis Tool to Explore \\Learned Representations in Transformers Models}

\author{
  Ben Hoover\\
  IBM Research\\
  MIT-IBM Lab\\
  \texttt{benjamin.hoover@ibm.com} \\
  \And
   Hendrik Strobelt \\
   IBM Research \\
  MIT-IBM Lab\\
   \texttt{hendrik.strobelt@ibm.com} \\
   \AND
   Sebastian Gehrmann \\
   Harvard SEAS \\
   \texttt{gehrmann@seas.harvard.edu} \\
}

\begin{document}

\maketitle

\begin{abstract}
Large language models can produce powerful contextual representations that lead to improvements across many NLP tasks. Since these models are typically guided by a sequence of learned self attention mechanisms and may comprise undesired inductive biases, it is paramount to be able to explore what the attention has learned. While static analyses of these models lead to targeted insights, interactive tools are more dynamic and can help humans better gain an intuition for the model-internal reasoning process. We present \textsc{exBERT}, an interactive tool named after the popular BERT language model, that provides insights into the meaning of the contextual representations by matching a human-specified input to similar contexts in a large annotated dataset. By aggregating the annotations of the matching similar contexts, \textsc{exBERT} helps intuitively explain what each attention-head has learned. 
\end{abstract}


\section{Introduction}

Neural networks based on the Transformer architecture have led to impressive improvements across many Natural Language Processing (NLP) tasks such as machine translation and text summarization~\citep{vaswani2017}. The Transformer is based on subsequent application of ``multi-head attention'' to route the model reasoning, and this technique's flexibility allows it to be pretrained on large corpora to generate contextual representations that can be used for other tasks.

Of these models, BERT is the most commonly used Transformer model for representation learning with several applications to transfer learning~\citep{devlin2018}. BERT and it's extensions~\citep{sanh2019distilbert,liu2019roberta} are dominating the standard language understanding benchmarks~\citep{wang2018glue,wang2019superglue}. Moreover, Transformer models have also had much success as autoregressively trained language models that can be used for generation tasks~\citep{radford2019language,keskar2019ctrl}.

It is not yet well-understood what information BERT encodes and how it uses the attention. To address this challenge, some research has focused on understanding whether BERT learns linguistic features such as Part Of Speech (POS), Dependency relationships (DEP), or Named Entity Recognition (NER)~\citep[e.g.,][]{tenney2018what, vigAnalysis2019, raganato2018, tenney2019}.
\citet{clark2019} found that heads at different layers learn specific linguistic structure despite being trained in a completely unsupervised manner, although many heads ostensibly learn redundancies. \citet{voita2019} further explore the nature of several specialized attention-heads to show that BERT depends on only a subset of the total heads and that overall model performance could be maintained when some heads were pruned. 

The above analyses provide an in-depth but static glimpse into the behavior of transformers. 
Experiments for these analyses are often supported by open-source repositories that implement the newest architectures and thus enable rapid experimentation~\citep{wolf2019transformers}. 
We similarly need flexible evaluation frameworks for Transformer models that allow the community to test hypotheses rapidly. 
Toward that end, visualization tools can offer concise summaries of useful information and allow interaction with large models. Attention visualizations such as BertViz by \citet{vigWorkshop2019} have taken large steps toward these goals by making exploration of BERT's attention fast and interactive for the user. However, interpreting attention patterns without understanding the attended-to embeddings, or relying on attention alone for a faithful interpretation, can lead to faulty interpretations~\citep{brunner2019, jain2019, wiegreffe2019}.

To address this challenge, we developed \textbf{\textsc{exBERT}}, a tool that combines the advantages of a robust but static analysis with a dynamic and intuitive view into both the attention and internal representations of the underlying model.\footnote{\textbf{exBERT} is available at \url{www.exbert.net}.}  
\textsc{exBERT} is agnostic to the underlying Transformer model and corpus and can thus be applied to different domains and languages. 
Similar to the static analysis by \citet{clark2019}, \textsc{exBERT} provides insights into both the attention and the token embeddings for the user-defined model and corpus by probing whether the representations capture metadata such as linguistic features or positional information.

\begin{figure}[t]
    \centering
    \includegraphics[width=\linewidth]{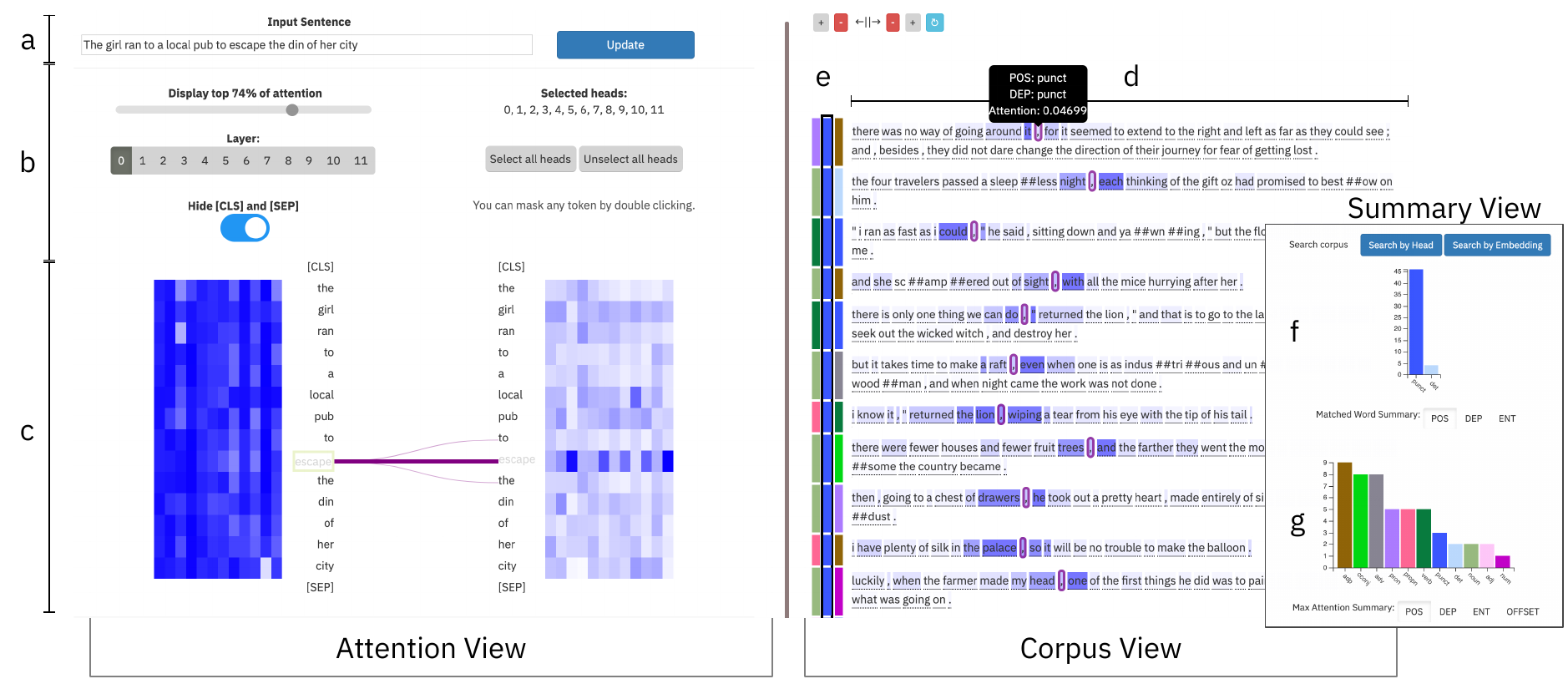}
    \caption{An overview of the different components of the tool. The token ``escape'' is selected and masked at 0-[all]. The results from a corpus search by token embedding are shown and summarized in (d-g). Users can enter a sentence in (a) and modify the attention view through selections in (b). Self attention is displayed in (c). The blue matrices show the attention of a head (column) to a token (row). Tokens and heads that are selected in (c) can be searched over the annotated corpus (shown: Wizard of Oz) with results presented in (d). Every token in (d) displays its linguistic metadata on hover. A colored summary of the matched token (black highlight) and its context is shown in (e), which can be expanded or collapsed with the buttons above it. The histograms in (f) and (g) summarize the metadata of the results in (d) for the matched token and the token of max attention, respectively.
    }
    \label{fig:over0}
\end{figure}

\section{Background}

\subsection{Transformer Models}

The Transformer model architecture as defined by \citet{vaswani2017} relies on multiple sequential applications \emph{self attention} layers. Self attention is the process by which each token within a sequence of inputs $Y$ computes attention weights over all other tokens in the same input. Within the process, the inputs are projected into a key, query, and value representation $W_k$, $W_q$, and $W_v$. The query and key representations are used to compute a weight for each token, which is then multiplied by that token's value, such that the values for one \emph{attention head} $h^{(i)}$ is defined as

\[ h^{(i)} =  \text{softmax} \Big( (Y W^{(i)}_q) ( Y W^{(i)}_k)^\top\Big) (Y W^{(i)}_v). \]

Transformer models typically use $n$ of these self attention heads in parallel. Their outputs $h^{(0)}, \ldots, h^{(n-1)}$ are concatenated and followed by a final linear projection. The output of this projection is used to calculate the token embedding used for the next layer.

\subsection{Transformer Analysis}
Previous analyses of transformer models focus on discovering how an unsupervised Transformer model learns to model our human understanding of language. For instance, \citet{clark2019} showed that individual heads seem to recognize common POS and DEP relationships,e.g., Objects of the Preposition (POBJ), Determinants (DET), and Possessive Adjectives (POSS), with high fidelity. \citet{vigAnalysis2019} also explored the dependency relations across heads and discovered that initial layers typically encode positional relations, middle layers capture the most dependency relations, and later layers look for unique patterns and structures. These insights are exposed visually and interactively through \textsc{exBERT}.

\section{Overview}

\textsc{exBERT} focuses on displaying a succinct view of both the attention and the internal representations of each token. The attention belonging to an input of length $N$ at a particular layer for a particular head can be understood as an $N \times N$ matrix, where each row represents the attention out of the corresponding token in the input, and each column represents the attention into that token. This is conducive to a representation of curves pointing from each token to every other token. Representations, on the other hand, are best understood by comparing the embedding of a token to the embeddings of other tokens in an annotated corpus. The most similar token embeddings, defined by a nearest neighbor search, can be viewed in their corpus' context in a language that humans can understand.

\subsection{Components}

Figure~\ref{fig:over0} shows an overview of the tool's three main components. The {\bf Attention View} provides an interactive view of the self attention of the model. Here, users can change layers, select heads, and view the aggregated attention. Tokens can be masked, and a selected token can be searched over the annotated corpus according to the methods laid out in Section~\ref{sec:search}. The results of this search are presented in the {\bf Corpus View}, with the highest-similarity matches shown first. The {\bf Summary View} shows histogram summaries of the matched metadata, which is useful for getting a snapshot of the metadata an embedding encodes in the searched corpus.

\subsection{Searching}
\label{sec:search}
Inspired by \citet{lstmVis2016,seq2seq2018}, \textsc{exBERT} performs a nearest neighbor search of embeddings on a reference corpus that is processed with linguistic features as follows. First, the corpus is split by sentence, and its tokens are labeled for desired metadata (e.g., POS, DEP, NER). 
Searching by token embeddings performs a Cosine Similarity (CS) search with the tokens in the corpus~\citep{faiss}. The top 50 matches are displayed and summarized for the user.  

Searching by head embeddings also involves a CS search against the corpus but requires an extension of the self attention definition. In our case, we define the \emph{head embedding} $E^{(l)}$ as

\begin{equation*}
    E^{(l)} = \textrm{Concat}(\normh{}{l,0}, \dots, \normh{}{l,n-1}),
\end{equation*}

where $\normh{}{l,i}$ is defined as the normalized representation of head $i$ at layer $l$.

This normalization makes it possible to perform a CS search over the head embeddings in our preprocessed corpus. To search the corpus for only a select subset $H_s \subseteq \{0, \dots, n-1\}$, we set all values of $\normh{}{l,i}$ to $0$ in our query head embedding $E^{(l)}$ where $i \notin H_s$.

\section{Case Study: BERT}

BERT is an instantiation of a Transformer model that can be used in applications that benefit from transfer learning~\citep{devlin2018}. 
BERT introduces special tokens into the typical training process in addition to the input tokens that are extracted using Byte-Pair Encoding (BPE)~\citep{BPE}. The architecture requires every input to start with a ``[CLS]'' and end with a ``[SEP]'' token and uses a technique called Masked Language Modeling (MLM) to develop its language model~\citep{devlin2018}. MLM works by replacing random tokens in the training corpus with ``[MASK]'' and training the model to determine what word should belong. Since there is no information in the original vector embedding of ``[MASK]'', BERT must rely on its internal representations to fill in the missing tokens. This provides an intuitive way to glimpse how BERT learns linguistic features in context. 

In the following cases, the reference corpus used is the Wizard of Oz,\footnote{\url{http://www.gutenberg.org/ebooks/55}} which is annotated and processed by BERT to allow for nearest neighbor searching. Whenever BPE tokenization splits a single word into multiple tokens, we assign its metadata to each component token. Special tokens like ``[CLS]'' and ``[SEP]'' have no linguistic features assigned to them, and are therefore removed from the reference corpus, which allows searches to always match a token that has intuitive meaning for users. This also allows the tool to apply the same corpus to different transformer models that may require different tokenization.

We now explore the layers and heads at which BERT learns the linguistic features of a masked token. We look at the following sentence:

\begin{center}
\emph{The girl ran to a local pub to escape the din of her city.}
\end{center}

Unless otherwise noted, the following examples are from the BERT$_{\textrm{base}}$ model, which has 12 layers and 12 heads per layer. We use the notation <layer>-<head> to refer to a single head at a single layer, and <layer>-[<heads>] to describe the cumulative attention of heads at a layer (e.g., 4-[0,3,9] to describe the sum of the attention of heads 0, 3, and 9 at layer 4). Note that we refer to layers and heads as 0 indexed in the tool.

\subsection{Behind the mask} \label{sec:btm}

We begin by masking the ``escape'' token in the example sentence at layer 1 and search what information is behind the ``[MASK]'' token's embedding. This setup is shown in Figure~\ref{fig:over0}. 
Note that at this early layer, the matching embeddings are most similar to punctuation (PUNCT) and determinants (DET), which are the most common tokens in English (Figure~\ref{fig:over0}f). Additionally, the maximum attention out of the MASKed token points to itself (Figure~\ref{fig:over0}c). We can observe that there is no meaningful linguistic information encoded in the mask's embedding at this layer.

\begin{figure}[t]
    \centering
    \includegraphics[width=\linewidth]{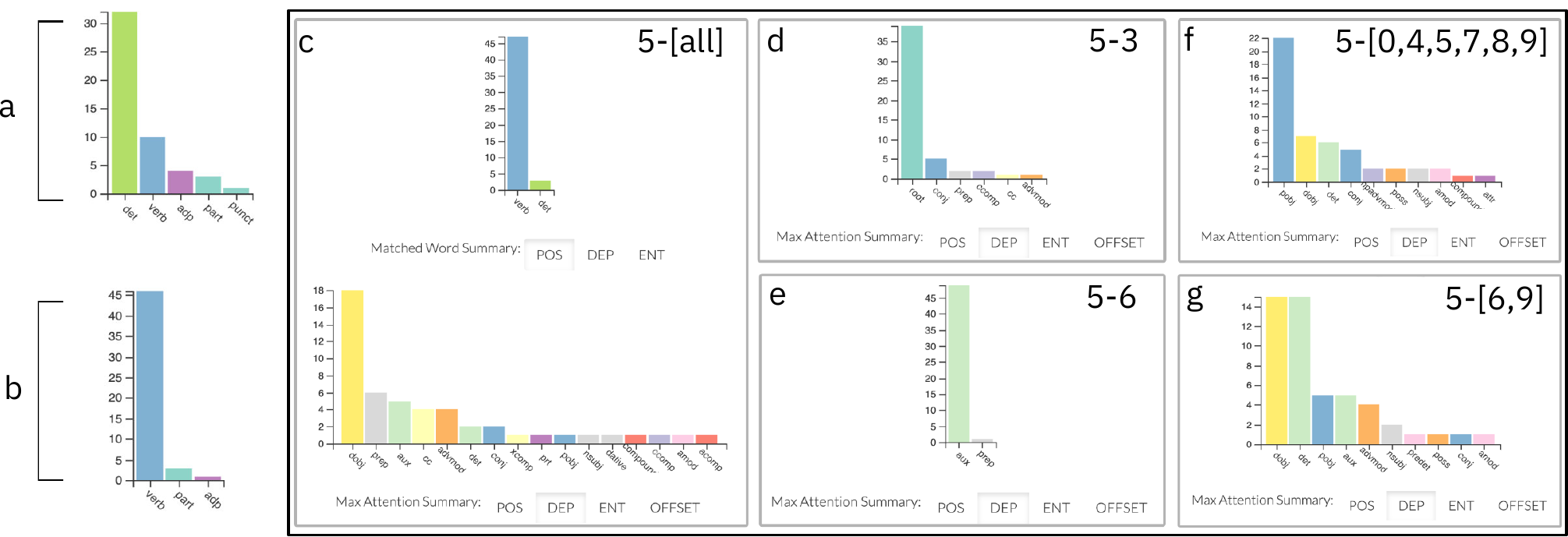}
    \caption{Left: searching by token embedding results. Histogram summaries shown at layers 5 (a) and 6 (b). Right: histogram summaries of searching by different head selections at layer 5.}
    \label{fig:histos}
\end{figure}

As layers progress, more POS information is added to the token embedding. The summarized layer 5 search results (Figure~\ref{fig:histos}a) show that we start to see some verb information creep into the embedding; however, it is not until layer 6 (Figure~\ref{fig:histos}b) that BERT is confident about the masked embedding being a verb. Subsequent layers seem to refine this estimation, with the embedding flirting with the possibility of the masked word being an adposition (ADP). At the final layer, BERT settles on the token being a VERB and wants to predict tokens such as ``pass'', ``mar'', ``see'', ``hear'', and ``breathe''. For the complete progression, see Figure~\ref{fig:supp1} in  Appendix~\ref{apdx:supp1}.

\subsection{Behind the heads}
Searching by the masked token's embedding helps show the information that is captured within the token itself, but it is useful to understand how the heads of the previous layer contributed to that information being encoded in the embedding. Going back to 5-[all], we see that the token embedding fails to embrace the masked ``escape'' token as a verb (Figure~\ref{fig:histos}a). However, a search by head embedding at that point reveals that BERT has already learned to attend to sentence structures where the most similar tokens in the corpus are verbs (\ref{fig:histos}c). Even at this early layer, it has learned to attend to the direct object (DOBJ) of that verb, a dependency that \citet{clark2019} showed was strongest in head 8-9. Exploring other individual heads at this point for DEP relationships reveals that 5-3 primarily detects the ROOT dependency (\ref{fig:histos}d) while 5-6 detects the AUX dependency (\ref{fig:histos}e).

This is useful, but it is not clear how all the heads were able to maximize their attention on the ``din'' token and thus detect the DOBJ pattern that was in 18 of the top 50 matches in the search. Inspecting all the heads, it is clear that no head \textit{individually} looks for DOBJ, and therefore that pattern must be detected through a combination of heads. Naively, we can strategically select the heads that maximize their attention to ``din'' (5-[0,4,5,7,8,9] shown in Figure~\ref{fig:histos}f), but find that these most normally find the object of the preposition (POBJ). Further exploration shows that the DOBJ pattern can be detected by 5-[6,9] (\ref{fig:histos}g), albeit with confusion to the DET dependency. It seems that complex dependencies like DOBJ can be detected in the early-middle layers of the model but rely on a combination of heads.

\subsection{More than position}

\begin{figure}[ht]
    \centering
    \includegraphics[width=\linewidth]{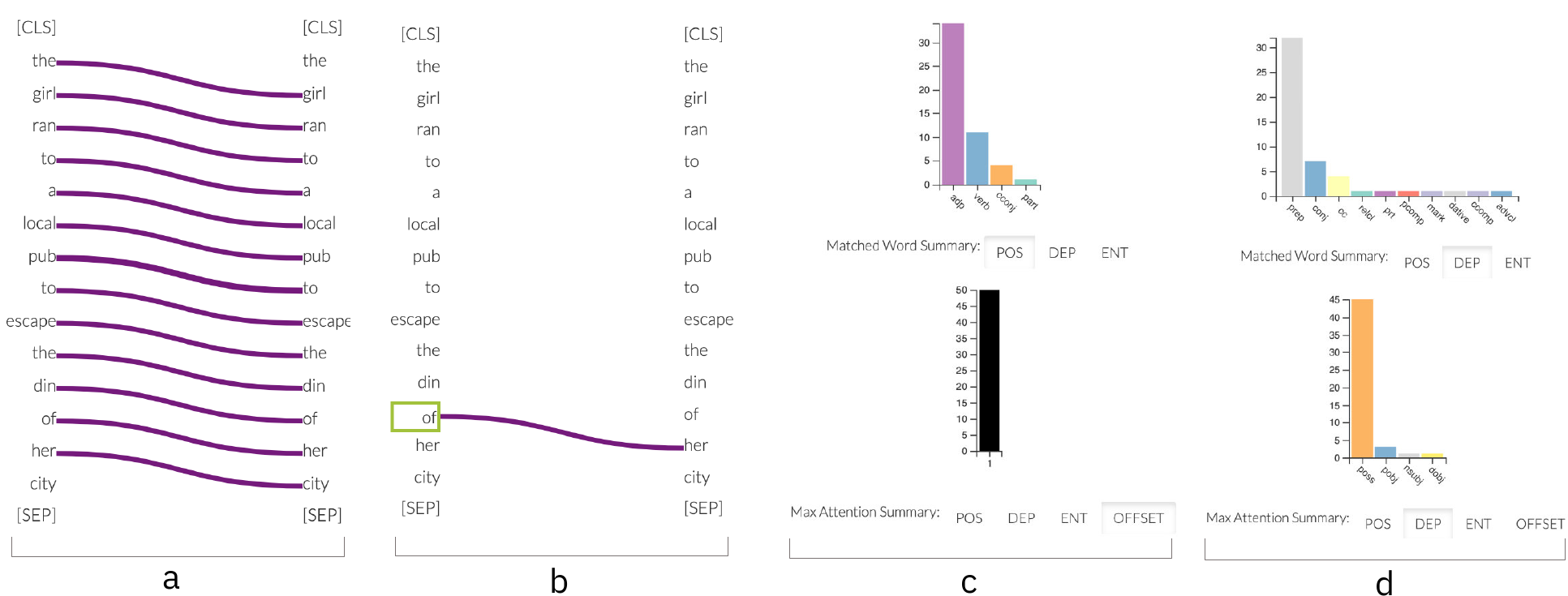}
    \caption{Exploration of positional heads, inspecting the positional head 2-0. The overview in (a) shows the head behavior always pointing to the following word, and the search token ``of'' is highlighted in (b). The matched results are summarized by POS and offset in (c) and by DEP in (d).}
    \label{fig:POL}
\end{figure}

Figure~\ref{fig:POL}a,c confirm that certain heads learn to attend to succeeding or preceding tokens. We call these heads \emph{positional heads} in that they detect an offset from the current token~\citep{clark2019}. Though simple, the positional head can encode important information about the attended-to word. Searching by head can reveal how much information from the token embeddings is visible to that head. A brief exploration of the attention in positional head 2-0 shows that the head is truly positional, matching the following word 50/50 times as seen in the lower histogram in Figure~\ref{fig:POL}c. It also seems to match the POS belonging to the seed token (in this case, ``of'' is an ADP).
The DEP summary at the bottom of Figure~\ref{fig:POL}d additionally shows that not only does the head match the POS of the seed token, but it has also learned to look for cases where the word following a preposition is possessive (e.g., him/her/its). This kind of exploration shows how much information the different attention heads see from the tokens they attend to.

\section{Conclusion}
In this paper, we have introduced an interactive visualization, \textsc{exBERT}, that uses linguistic annotations, interactive masking, and nearest neighbor search to help revealing an intelligible structure about learned representations in transformer models.
We demonstrate the applicability of \textsc{exBERT} to a specific case study for a BERT model across the Wizard of Oz corpus.
The source code and demo are available at \url{www.exbert.net}, providing the community an opportunity to rapidly experiment with learned Transformer representations and gain a better understanding of what these models learn.

\section{Acknowledgements}
We thank Jesse Vig for his helpful feedback.

\bibliography{refs}

\clearpage
\appendix


\clearpage
\section{Embedding results across layers}
\label{apdx:supp1}

\begin{figure}[h]
    \centering
    \includegraphics[width=0.8\linewidth]{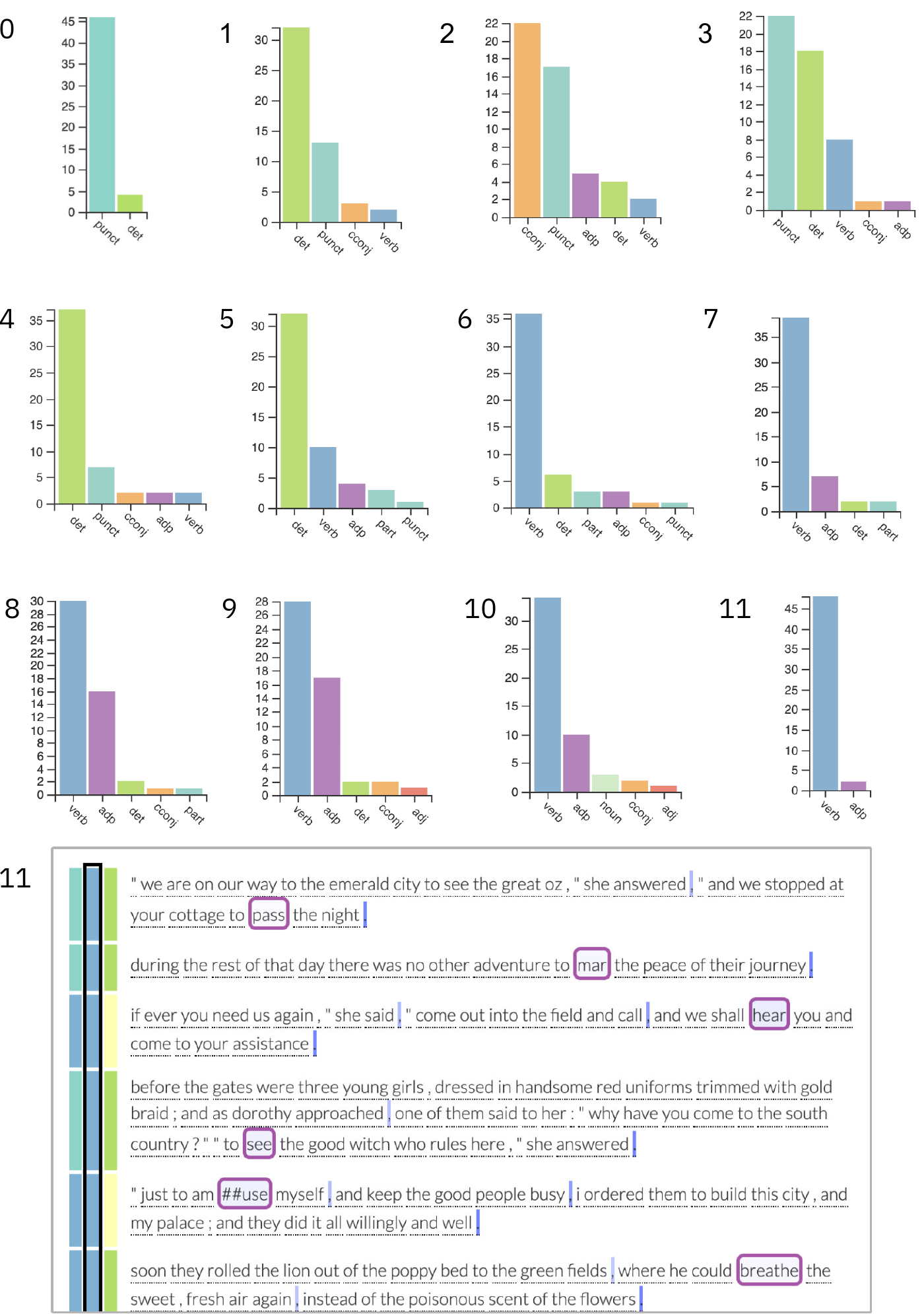}
    \caption{Token embeddings for the ``escape'' token setup in \ref{sec:btm} across every layer. The matched tokens at the output of the model are shown in the bottom corpus inspector view, whereas summaries are shown for all the other layers}
    \label{fig:supp1}
\end{figure}

\end{document}